\documentclass[letterpaper]{article} 
\pdfoutput=1
\usepackage{aaai25}  
\usepackage{times}  
\usepackage{helvet}  
\usepackage{courier}  
\usepackage[hyphens]{url}  
\usepackage{graphicx} 
\usepackage{natbib}  
\usepackage{caption} 
\usepackage{algorithm}
\usepackage{algorithmic}
\usepackage{amsmath}
\usepackage{amssymb}
\usepackage{comment}
\usepackage{array}
\usepackage{newfloat}
\usepackage{listings}
\DeclareCaptionStyle{ruled}{labelfont=normalfont,labelsep=colon,strut=off} 
\floatstyle{ruled}
\newfloat{listing}{tb}{lst}{}
\floatname{listing}{Listing}

\title{Quality over Quantity: Boosting Data Efficiency Through \\Ensembled Multimodal Data Curation}
\author{
    Jinda Xu\textsuperscript{\rm 1}\equalcontrib,
    Yuhao Song\textsuperscript{\rm 2}\equalcontrib,
    Daming Wang\textsuperscript{\rm 2},
    Weiwei Zhao\textsuperscript{\rm 1},
    Minghua Chen\textsuperscript{\rm 2},\\
    Kangliang Chen\textsuperscript{\rm 2}\thanks{Project leader.},
    Qinya Li\textsuperscript{\rm 1}\thanks{Corresponding author.}
}
\affiliations{
    \textsuperscript{\rm 1}Shanghai Key Laboratory of Scalable Computing and Systems, Shanghai Jiao Tong University\\
    \textsuperscript{\rm 2}HAOMO.AI Technology Co., Ltd.\\

    daidan2502@sjtu.edu.cn, \{songyuhao, wangdaming\}@haomo.ai, zhaoweiwei@sjtu.edu.cn, \{chenminghua, chenkangliang\}@haomo.ai, qinyali@sjtu.edu.cn

}

\begin{document}
\maketitle

\begin{abstract}

In an era overwhelmed by vast amounts of data, the effective curation of web-crawl datasets is essential for optimizing model performance. This paper tackles the challenges associated with the unstructured and heterogeneous nature of such datasets. Traditional heuristic curation methods often inadequately capture complex features, resulting in biases and the exclusion of relevant data. We introduce an advanced, learning-driven approach, \textbf{E}nsemble \textbf{C}uration \textbf{O}f \textbf{DA}ta \textbf{T}hro\textbf{U}gh \textbf{M}ultimodal Operators (EcoDatum), incorporating a novel quality-guided deduplication method to ensure balanced feature distributions. EcoDatum strategically integrates various unimodal and multimodal data curation operators within a weak supervision ensemble framework, utilizing automated optimization to score each data point effectively. EcoDatum, which significantly improves the data curation quality and efficiency, outperforms existing state-of-the-art (SOTA) techniques, ranked 1\textsuperscript{st} on the DataComp leaderboard, with an average performance score of 0.182 across 38 diverse evaluation datasets. This represents a 28\% improvement over the DataComp baseline method, demonstrating its effectiveness in improving dataset curation and model training efficiency. Our code is available at \url{https://github.com/Daming-W/EcoDatum}.

\end{abstract}

\section{Introduction}

The vast amount of data presents opportunities for training advanced deep learning models, but it also introduces significant noise and irrelevant information, which can hinder model effectiveness. In both academia and industry, the need for robust data curation techniques to extract meaningful signals from extensive digital information has become a pressing concern. In web-crawled datasets, data curation is a multi-faceted task involving various stages and methodologies. The core objective is to identify and retain high-quality samples while discarding noise or mitigating the impact of irrelevant data \cite{hoffmann2022training}. This process is crucial for optimizing model performance in the deep learning framework.

\begin{figure}
    \centering
    \includegraphics[width=1\linewidth]{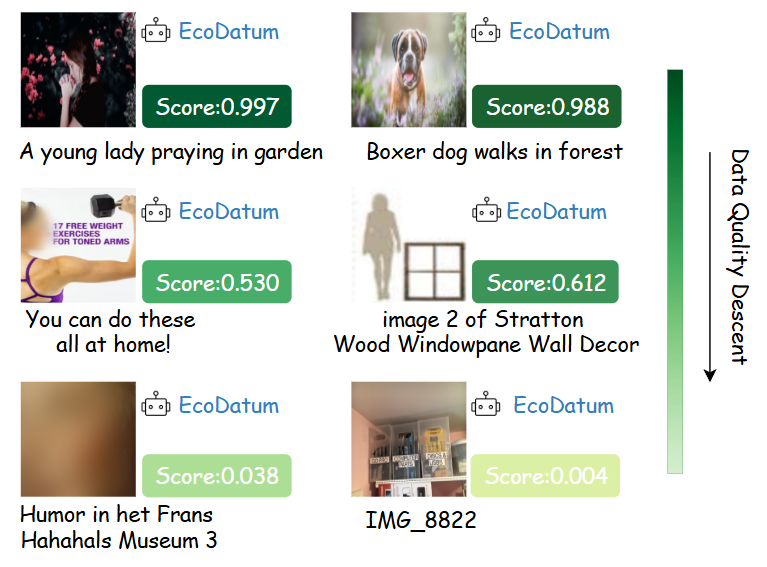}
    \caption{Web-crawled image-text datasets often vary in quality. EcoDatum addresses this by integrating unimodal and multimodal operators to discriminate data quality, guiding the selection process to enhance data curation efficiency.}
    \label{fig:introdemo}
\end{figure}

Web-crawled data is inherently unstructured, diverse, and constantly evolving, making it essential to develop adaptive curation methods capable of handling such complexity\cite{hoffmann2022training}. Traditionally, data curation methods have relied heavily on heuristic filtering approaches based on manually defined content attributes, such as image resolution, graphic geometry, textual length, and linguistic complexity. While these heuristic methods provide a basic means of recognizing low-quality samples, they fail to adequately capture the subtle features of web-crawled data and may introduce biases or overlook relevant information \cite{mindermann2022prioritized, maini2021dataset}. To address these limitations, researchers are increasingly adopting automated curation methods that leverage deep learning techniques, including natural language processing, computer vision and cross-modal representation learning, to achieve a balance of quality and quantity.\cite{schuhmann2021laion, brown2020language, torkashvand2023deep}. 

This research proposes a data curation framework called EcoDatum to address the aforementioned issues. Specifically, we implement a range of efficient data curation strategies as operators, to enhance the data curation process and to achieve cross-modal data alignment at various levels of granularity, as detailed in Figure \ref{fig:introdemo}. However, a simple combination of these operators may introduce bias and lead to insufficient utilization of their individual strengths. To fully capture their synergies, we develop a weak supervision ensemble framework that integrates these operators, achieving a synergistic effect. Furthermore, to enhance the integration efficiency of unimodal and multimodal operators, EcoDatum introduces an automated optimization approach. This is achieved by tuning the weak supervision integration module using a composite metric and a tiny-labeled dataset.

As a novel weak supervision-based framework for multimodal data curation, EcoDatum achieves state-of-the-art performance on the DataComp data filtering track\cite{gadre2024datacomp}. The visual language model trained on curated data demonstrates outstanding results over 38 downstream tasks, highlighting its strong generalizability. Extended experiments demonstrate the effectiveness of this research in understanding various operators in cross-modal data management, offering insights for future work.

Our main contributions are as follows:

\begin{enumerate}
    \item We propose an auto-optimized ensemble framework, EcoDatum, which integrates techniques to enhance data quality and curate multimodal data, ensuring aligned, high-quality inputs for visual language pretraining.
    
    \item We introduce a search-based optimization algorithm for weak supervision labeling function tuning that enhances the curation process and boosts the system's robustness.

    \item EcoDatum surpasses existing state-of-the-art techniques in the DataComp benchmark over 38 downstream tasks and ranks 1\textsuperscript{st} on the leaderboard\footnotemark.
\end{enumerate}

\footnotetext{https://www.datacomp.ai/dcclip/leaderboard.html}

\section{Related Work}
\label{sec:headings}

\subsection{Data Curation for Web-crawled Datasets}

Recent research underscores the critical role of data curation in enhancing model performance with large-scale image-text datasets. Various studies focus on improving dataset quality through curation methods, such as enhancing the descriptiveness and cross-modal feature alignment of image-text pairs, and reducing redundancy \cite{radenovic2023filtering, nguyen2024improving, abbas2023semdedup}.

In a broader context, DataComp is a benchmark designed to evaluate the performance of multimodal models on large-scale, real-world datasets\cite{gadre2023datacomp}. Recent advancements in the DataComp benchmark highlight notable progress in data curation techniques. Yokoo et al.\cite{yokoo2023leveraging} advanced data filtering using image-text similarity and caption modification, achieving notable progress in the Filtering and Bring Your Own Data (BYOD) Tracks. Yu et al.\cite{yu2023devil} evaluated data selection criteria's impact on model performance, while Chen et al.\cite{chen2024data} introduced DataJuicer for managing large-scale datasets. Nguyen et al.\cite{nguyen2024improving} enhanced image captioning for better modality alignment, and Maini et al.\cite{maini2023t} presented T-MARS for improved visual representation by bypassing text feature learning. Additionally, significant contributions have been made in the areas of synthetic data, contrastive learning, image-text alignment, and few-shot learning\cite{chen2020simple, radford2021learning, jia2021scaling, li2022blip, alayrac2022flamingo}.

Despite the significant advancements in data curation achieved by Radenovic et al.\cite{radenovic2023filtering} and Nguyen et al.\cite{nguyen2024improving} through enhancing data relevance, existing automated filtering methods may still exclude valuable but less conventional data points or introduce biases by focusing too narrowly on specific aspects, potentially overlooking broader contextual information. 

\subsection{Ensemble Learning}
Ensemble learning, which combines multiple models to improve performance and generalization, has long been a foundational approach in machine learning. Classic methods such as Bagging\cite{breiman1996bagging} and Boosting \cite{freund1997decision} initially demonstrated how model aggregation could reduce variance and improve accuracy.

Afterwards, ensemble learning has been increasingly applied to specialized tasks. Zimek, Schubert, \& Kriegel \cite{zimek2012survey}  highlighted how ensemble methods enhance outlier detection by aggregating results from multiple models. Beluch et al.\cite{beluch2018power} demonstrated that ensemble-based uncertainty sampling can significantly improve efficiency in active learning. Rasmus et al. \cite{rasmus2015semi} demonstrated that ensemble techniques enhance semi-supervised learning by effectively utilizing both labeled and unlabeled data. Song et al.\cite{song2022learning} used ensemble methods to improve data quality by detecting and filtering noisy or mislabeled data.

These advancements illustrate how ensemble techniques refine data preprocessing and improve model inputs.

\section{Method}

\subsection{Overview}

\begin{figure*}[ht]
    \centering
    \includegraphics[width=1\linewidth]{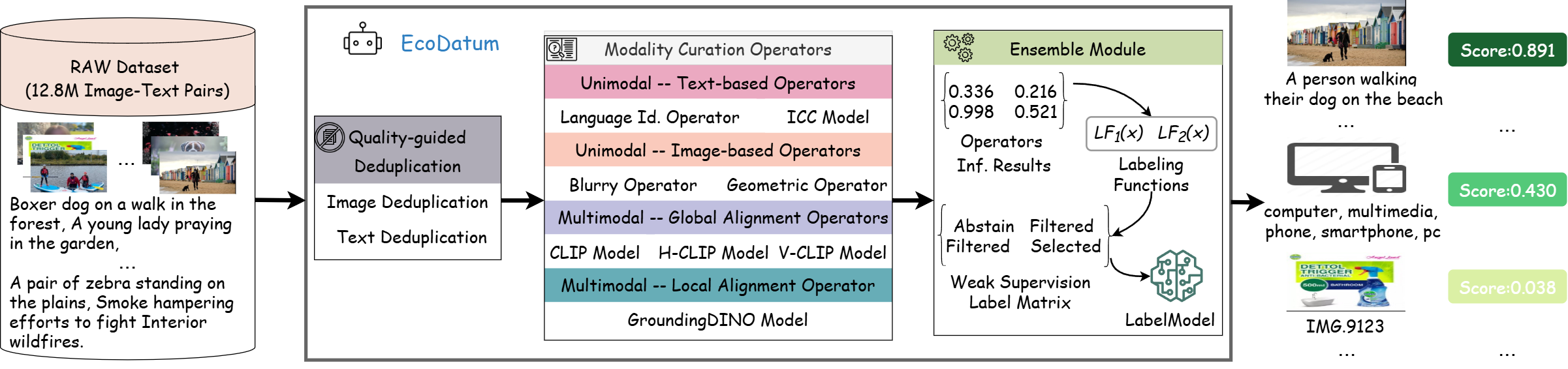}
    \caption{Overview of the EcoDatum Framework. EcoDatum utilizes quality-guided deduplication along with an ensemble of unimodal and multimodal data curation operators, that strategically curate multimodal datasets. This integrated approach systematically scores each data point, ensuring optimal quality and alignment for effective visual-language pretraining.}
    \label{fig:EcoDatum}
\end{figure*}

As illustrated in Figure \ref{fig:EcoDatum}, EcoDatum enhances the pretraining effectiveness of multimodal models like CLIP\cite{radford2021learning} by strategically selecting high-quality subset \( \hat{S} \) from the original dataset \( S \). This targeted data curation improves the model's zero-shot performance on diverse tasks. The framework utilizes an ensemble of specialized data operators for comprehensive quality assessment, which addresses various dimensions including image filtering, text analysis, and cross-modal alignment at multiple granular levels. Automated optimization enables the weak supervision system to generate quality scores for data samples, thus minimizing manual input and enhancing the precision of threshold settings. Consequently, EcoDatum streamlines the data curation process and significantly elevates the quality, ensuring the dataset meets the rigorous requirements for model training.

\subsection{Quality Guided Deduplication}

\begin{figure}[!h]
    \centering
    \includegraphics[width=1.0\linewidth]{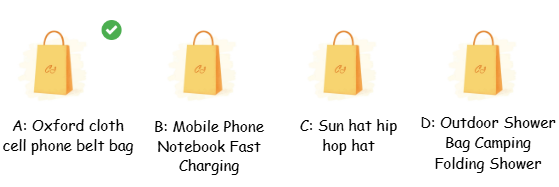}
    \caption{Quality Guided Deduplication retains the samples with better cross-modal alignment in duplicate groups to enhance the overall quality and achieve optimal data distributions.}
    \label{fig:dedup}
\end{figure}

To improve our dataset's diversity and distribution, we employ a quality-guided deduplication process that removes redundant text-image pairs. This approach uses perceptual hashing \cite{farid2021overview} to generate hash codes, identifying duplicates based on visual and textual content. Subsequently, the CLIP model assesses the semantic coherence of each duplicate group, allowing us to retain text-image pairs with the highest CLIP scores, as shown in Figure \ref{fig:dedup}. This selective retention enhances the dataset by preserving the most relevant and semantically rich examples, minimizing redundancy while maintaining quality and diversity.

\vspace{-0.2cm}
\subsection{Unimodal and Multimodal Curation Operators}  
EcoDatum enhances the quality of multimodal datasets by implementing rigorous unimodal and multimodal curation operators. The unimodal curation operators systematically filter out low-quality visuals and evaluate textual data for concreteness and relevance using both language identification and Image Caption Concreteness (ICC) metric \cite{yanuka2024icc}. Multimodal curation integrates these approaches with advanced alignment techniques, employing models like GroundingDINO, an advanced open-set object detector \cite{liu2023grounding} for precise local feature alignment and the CLIP model, for global semantic coherence. Together, these strategies ensure the curated dataset is of high quality, with well-aligned multimodal content.

\subsubsection{Unimodal Curation Operators.}
For images, the specific heuristic operators filter out blurred and low-quality visuals. For texts, the FastText \cite{joulin2016fasttext} model identifies the language and the ICC metric evaluates the relevance and clarity of textual data using a pre-trained autoencoder. 

\textit{Image-based quality filtering.} 
Low-quality images can severely impact the learning of visual semantics. Our unimodal operators, based on heuristic rules, enhance dataset quality by filtering out images with detrimental attributes. The Geometric Operator targets images with non-standard aspect ratios that distort geometric relationships and compromise visual integrity when resized. Additionally, the DataComp dataset contains many intentionally blurred images to meet privacy standards, which reduces the visual detail crucial for effective model training. The Blurry Operator identifies and removes these excessively blurred images, ensuring that the curated dataset retains high visual quality.

\textit{Text-based caption assessment.} 
We leverage the FastText model to identify and remove captions in rare languages, enhancing the linguistic consistency of our dataset. Additionally, we use the ICC metric, developed by a pre-trained autoencoder, to independently assess and filter captions. EcoDatum ensures the dataset retains only concrete and relevant captions, directly corresponding to their images.

\subsubsection{Multimodal Curation Operators.}  

\begin{figure}[]
    \centering
    \includegraphics[width=1\linewidth]{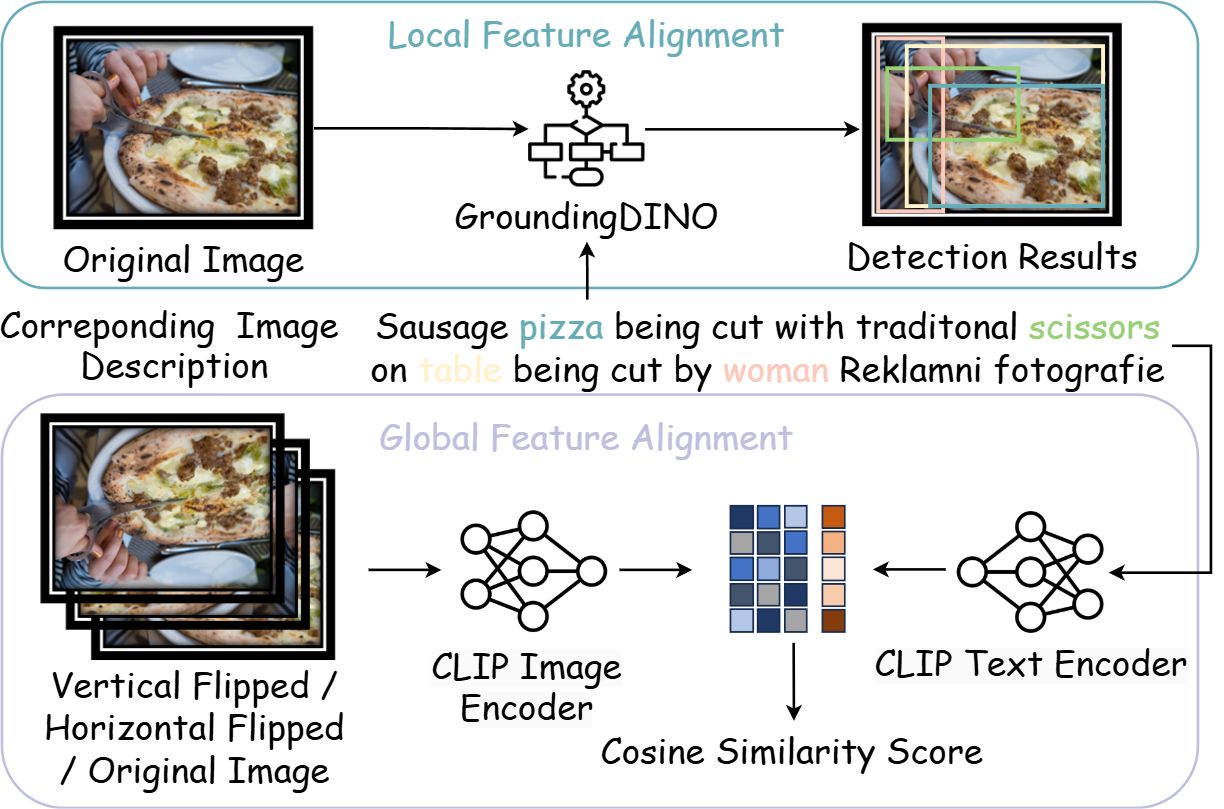}
    \caption{Illustration of multimodal curation operators integrating local and global cross-modal feature alignments. }
    \label{fig:globallocal}
\end{figure}

EcoDatum enhances multimodal data curation by integrating both global and local image-text features, as shown in Figure \ref{fig:globallocal}. We employ GroundingDINO for precise local feature alignment, ensuring detailed correspondence between text and images at the object level. Additionally, we utilize the CLIP model, augmented with innovative adaptations, to maintain global semantic coherence throughout the dataset.

\textit{Local Cross-Modal Feature Alignment.} 
We utilize GroundingDINO for the precise alignment of text descriptions with corresponding visual content. It integrates and analyzes text and visual data, effectively identifying relevant phrases in captions and accurately localizing associated visual elements within images, ensuring precise text-to-image alignment without prompt modification.

To quantitatively assess the alignment between text and images, we develop a metric based on the count of bounding boxes with confidence scores exceeding a predefined threshold, as shown in Eq (\ref{eq:gdino}). This metric serves to highlight the degree of correspondence between textual descriptions and visual representations. A higher count of accurate detections indicates richer, more detailed scenes, signifying that these data points are of higher value for training and subsequent applications. Data points that do not meet this threshold can be effectively filtered out, including those where the described objects do not visually correspond to the images or where the textual descriptions are insufficiently specific. This ensures our dataset excludes mismatches and generalities, retaining only high-quality, relevant multimodal content.

\begin{equation}
\label{eq:gdino}
    Count_{\text{GroundingDINO}} = \sum_{i=1}^{n} \mathbf{\{x_i > \text{t}\}}
\end{equation}
\noindent
where \( x_i \) represents the confidence score of the \( i \)-th detected object, \( n \) represents the total number of objects detected, and \( t \) represents the predefined threshold.

This operator enhances the ability to curate multimodal data effectively, ensuring that the dataset maintains the most relevant and accurately aligned text-image pairs locally.

\textit{Global Cross-Modal Feature Alignment.} 
In this module, EcoDatum utilizes the CLIP model, celebrated for its ability to assess the global semantic similarity between text descriptions and their visual counterparts. However, the effectiveness of the CLIP-Score can be compromised when images contain textual content that overlaps with captions. This issue is observed in 40\% of the LAION dataset\cite{schuhmann2022laion} and 20\% of the Datacomp dataset \cite{maini2023t}. To mitigate this, we implement an innovative adaptation known as Flip-CLIP, which includes Horizontal-CLIP (H-CLIP) and Vertical-CLIP (V-CLIP) techniques, inspired by \cite{yu2023devil}. Before computing the CLIP scores, images are flipped horizontally or vertically, reducing the model's bias towards text-based features and enabling more equitable evaluations of purely visual elements. The development of Flip-CLIP is motivated by the observation that OCR tasks often disproportionately influence the standard CLIP score, especially when the image-text is overlapped.

By integrating both CLIP-Score and Flip-CLIP-Score, we foster the model's ability to learn from visual content independently of textual influences, thereby enhancing EcoDatum's capability to process and understand global visual features without excessive bias towards textual elements.

\subsection{Modality Operators Ensemble}

Given the vast volume of data and the high cost associated with obtaining high-quality labeled data, the availability of reliable labels is often limited. EcoDatum introduces a weak supervision labeling system that allows the efficient generation of quality-indicated labels at scale, mitigating the challenges of data scarcity and ensuring a more robust data quality assessment. 
In this study, data curation is abstracted as a data quality discrimination task, aiming to identify ``high-quality'' data. This ensemble-based system further enhances the capabilities of the data operators described above.

Specifically, EcoDatum employs a weak supervision ensemble model called LabelModel \cite{ratner2017snorkel, bach2019snorkel} into the scope of data curation research, which integrates signal sources abstracted from unimodal and multimodal operators for data quality evaluation. This integration balances the limitations of individual operators and significantly reduces their erroneous impacts.

Each operator serves as an independent weak supervision signal source, assessing data quality from its unique dimension.
The integration approach in this work uses LabelModel to combine multiple operators, automatically inferring a data quality score for each data sample by modeling the accuracy and relationships of these operators.

This process begins by matching each operator with corresponding labeling functions (LFs)~\cite{ratner2017snorkel}, which converts the operator's inferred score \( s \) of the data sample \( x_i \) into weak supervision label \( L \), as shown in Eq (\ref{eq:lf_transform}). The LFs compute operators' inference results with the mean value \( b \) and the standard deviation \( \beta \)  of the decision boundary to transform continuous scores into discrete labels. These labels are then aggregated to form a comprehensive weak supervision label matrix. In this context, weak supervision labeling with ``Abstain'' addresses situations where LFs face unclear features or inapplicable rules. Allowing the LabelModel to abstain from assigning labels in these cases prevents the generation of incorrect labels. This approach enhances the LabelModel's ability to integrate diverse LFs by learning transformed matrix, particularly when they exhibit different biases and error patterns, thereby increasing the model's robustness when handling heterogeneous data.

\begin{equation}
\label{eq:lf_transform}
L_{x_ij} = 
\begin{cases}
1, & \text{if } s_{x_ij} \geq b_j + \beta_j \quad \text{(Selected)} \\
0, & \text{if } s_{x_ij} \leq b_j - \beta_j \quad \text{(Filtered)} \\
-1, & \text{if } b_j - \beta_j < s_{x_ij} < b_j + \beta_j \quad \text{(Abstain)}
\end{cases}
\end{equation}

\begin{figure*}[!h]
    \centering
    \includegraphics[width=0.98\linewidth]{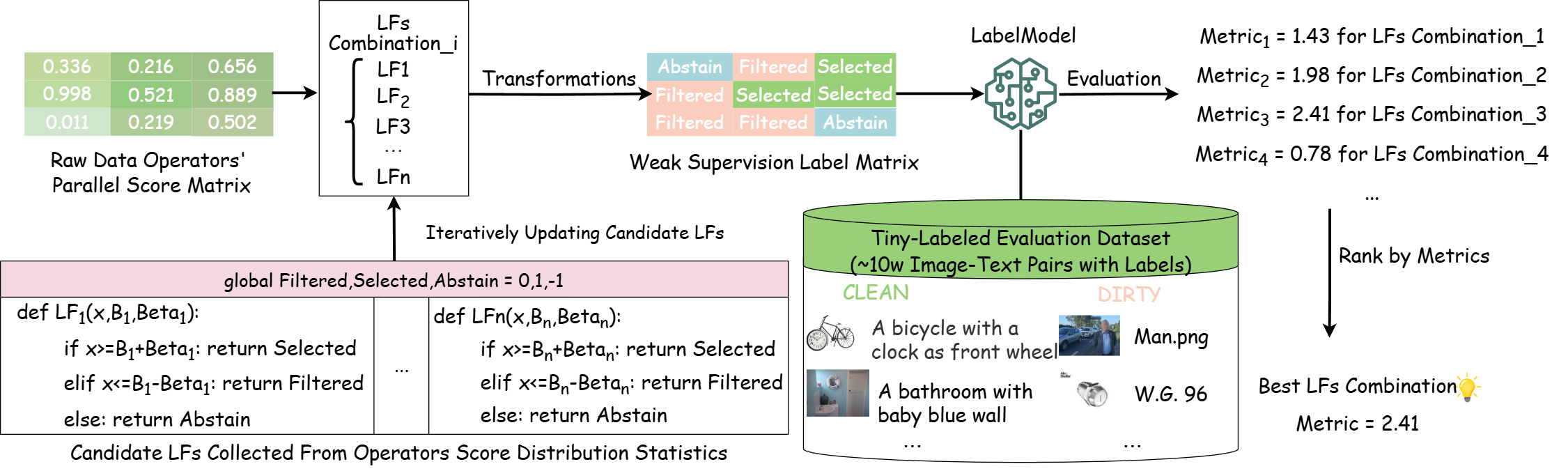}
    \caption{Overview of Search-Based LFs Combinations Optimization. This method optimizes LFs to create a more accurate weak supervision label matrix. The approach involves automatically searching optimal LFs to improve the matrix used by the LabelModel, evaluating LF combinations using a specialized composite metric over a gathered tiny-labeled dataset. This process boosts the LabelModel’s ability to assess data quality by refining operator interrelations and minimizing manual adjustments.}
    \label{fig:search}
\end{figure*}

The LabelModel learns the transformed weak supervision label matrix \( L_M \), estimating the weight \( w_j \) for each LF. These weights are used to combine the outputs of all LFs, ultimately generating a score for each data sample, which determines whether it is retained or filtered out.
This approach enhances the comprehensiveness and robustness of data quality evaluation, which ultimately allows the LabelModel to score all raw data and reflect quality.

\subsection{Search-based Optimization}

A novel search-based optimization method is introduced to enhance the design of LFs, improving the generation of a more accurate weak supervision label matrix for LabelModel modeling, as shown in Figure \ref{fig:search}. This method addresses the challenge of converting operator-derived scores into labels by automatically optimizing LFs, reducing the need for manual experimentation. To further optimize the performance of the ensemble, EcoDatum proposes a composite metric that integrates the LabelModel’s data quality assessment capability with the attributes of LFs combination from the transformation steps, enabling a refined weak supervision label matrix. This approach enhances the LabelModel's ability to analyze operator interrelations and importance, producing quality scores for data samples that closely approximate the ideal.

The evaluation stage automatically constructs a small labeled dataset containing ``clean" and ``noisy" samples. Clean data, labeled ``1", are sourced from the COCO dataset\cite{lin2014microsoft}, while ``noisy" samples, labeled ``0", are randomly sampled from the DataComp dataset to introduce both unimodal and multimodal noise and include added cross-modal noise through image-text pair exchanges. This setup tests the LabelModel’s ability to differentiate data quality via the F1-tiny scores in Eq (3). Importantly, this dataset is only used for assessing the LabelModel's performance and does not contribute to training the model or optimizing Eq (3) coefficients, ensuring unbiased validation of the LF effectiveness.

\begin{algorithm}[!t]
\caption{Search Optimization for LFs}
\label{alg:search_optimization}
\begin{algorithmic}[1]
\STATE \textbf{Input:} 
\STATE \quad \parbox[t]{\dimexpr\linewidth-\algorithmicindent}{
    Raw dataset $D_{raw}$, Tiny-labeled dataset $D_{tiny}$,  \\
    Operator $Op$,     Evaluation metrics $M$, \\
    Candidate LF Combinations $LFs\space Combs$, 
}
\STATE \textbf{Output:} 
\STATE \quad Optimal $LFs\space Comb$
\STATE Initialize $M^*$

\FOR{each $LFs \in LFs\space Combs$}
    \STATE Convert weak supervision label matrix $L_M$ from $LFs$ 
    \STATE Train LabelModel on $L_M$
    \STATE Predict and evaluate LabelModel on $D_{tiny}$
    
    \IF{$M(LFs) > M^*$}
        \STATE Update $M^* \gets M(LFs)$ and $LFs^* \gets LFs$
    \ENDIF
\ENDFOR

\STATE \textbf{return} $LFs^*$
\end{algorithmic}
\end{algorithm}

To evaluate the data quality discrimination capacity of the LabelModel after learning generated weak supervision label matrics with different combinations of LFs, this research develops a specialized composite metric, shown in Eq (\ref{eq:lm_metric}), which combines classification metrics against ground truth and further incorporates the attributes of each operators' LFs, specifically measuring the $f_{\text{Overlap}}$, $f_{\text{Conflict}}$, and $f_{\text{Coverage}}$. Here, they respectively indicate the frequency of agreement among LFs, the extent of disagreements, and the proportion of data labeled by at least one function.

\begin{equation}
\label{eq:lm_metric}
M = \alpha_1 \cdot F1_{\text{tiny}} + \alpha_2 \cdot f_{\text{Overlap}} - \alpha_3 \cdot f_{\text{Conflict}} + \alpha_4 \cdot f_{\text{Coverage}} 
\end{equation}

\noindent Here, \(\alpha_1\), \(\alpha_2\), \(\alpha_3\), \(\alpha_4\) are coefficients that are determined through a few rounds of experiments. These coefficients are tuned to optimize the balance between classification performance on the tiny labeled dataset and the contributions from overlap, conflict, and coverage metrics within the weak supervision labeling framework.

\begin{equation}
\label{eq:overlap}
f_\text{Overlap} = \frac{1}{n} \sum_{i=1}^{n} \mathbb{I}\left(\sum_{j=1}^{m} \mathbb{I}(LF_j(x_i) \neq 0) > 1\right)
\end{equation}

\vspace{-0.1cm}

\begin{equation}
\label{eq:conflict}
f_\text{Conflict} = \frac{1}{n} \sum_{i=1}^{n} \mathbb{I}\left(\exists \, j_1 \neq j_2 {, } LF_{j_1}(x_i) \neq LF_{j_2}(x_i) \neq 0 \right)
\end{equation}

\begin{equation}
\label{eq:coverage}
f_\text{Coverage} = \frac{1}{n} \sum_{i=1}^{n} \mathbb{I}\left(\sum_{j=1}^{m} \mathbb{I}(LF_j(x_i) \neq 0) \geq 1\right)
\end{equation}

These metrics capture the model's effectiveness in integrating weak supervision signals. For example, even if the model achieves a high F1-score \(F1_{tiny}\) on a tiny labeled dataset, significant conflicts between LFs or low coverage might still lead to instability in real-world applications. By optimizing these composite metrics, we can enhance the model’s generalization ability across different datasets.

The overall optimization process involves repeatedly constructing the LabelModel across candidate LFs combinations and using the results of the aforementioned composite metrics to identify the optimal LabelModel leaned by each weak supervision label matrix, as shown in Algorithm \ref{alg:search_optimization}. This optimized model is then applied to the final data curation task. By employing the optimized LabelModel, the overall framework can maximize the robustness of data quality assessment, balance the limitations of individual operators, and ultimately enhance overall data efficiency.

\section{Experiments}
\subsection{Dataset and Benchmark}

The DataComp benchmark uniquely emphasizes data curation over model development. Unlike typical machine learning competitions that seek the best model with a fixed dataset, DataComp challenges participants to curate optimal datasets using fixed training code. This highlights the crucial role of high-quality, well-curated data in enhancing model performance. We choose the small-scale filtering track to validate the proposed framework EcoDatum, we curate a subset from a vast pool of 12.8 million image-text pairs from Common Crawl, adhering to the competition's constraints of fixed training parameters and computational budgets. Our objective is to efficiently filter this dataset, ensuring consistency in training iterations regardless of dataset size.

The effectiveness of our curated dataset is evaluated across 38 diverse datasets, including ImageNet, 6 ImageNet distribution shift datasets, 12 datasets from the Visual Task Adaptation Benchmark, three retrieval datasets, and several others\cite{gadre2024datacomp}. This extensive range of evaluation datasets tested the generalizability and robustness of EcoDatum, providing a comprehensive assessment of their impact on model training across various real-world scenarios.

\subsection{Implementation Details}
For the local cross-modal curation operator, we employ the GroundingDINO-based model\cite{liu2025grounding} with Swin-Large as the image backbone and BERT-Base \cite{devlin2018bert} for encoding text, setting confidence thresholds at 0.1 to retain more potentially feature-aligned data. In global cross-modal curation, we use the CLIP-ViT-Large-14 architecture\cite{radford2021learning}. In determining final data volume, we conducted extensive experiments and reviewed related works, concluding that approximately the top 40\% samples by the EcoDatum quality score after deduplication (around 3.5M) provide the best balance between quality and quantity. Experiments utilized 8 NVIDIA A100 GPUs, The training and evaluation process required 2.5 hours. Data curation for the 12.8 million dataset involved approximately 10 hours.

\subsection{Result Analysis}
\begin{table*}[!ht]
\centering
\renewcommand{\arraystretch}{1.1}
\setlength{\tabcolsep}{6pt} 
\begin{tabular*}{\textwidth}{@{\extracolsep{\fill}}lccccccccc}
\hline
 & \textbf{No Filtering} & \textbf{LAION} & \textbf{Datacomp} & \textbf{CLIP} & \textbf{HYPE} & \textbf{LINE} & \textbf{T-MARS} & \textbf{WS} & \textbf{Ours} \\
\hline
\textbf{Dataset Size} & 12.8M & 1.3M & 3M & 3.8M & 2.3M & 4.5M & 2.3M & 4.1M & 3.5M \\
\textbf{Avg. Perf.} & 0.132 & 0.133 & 0.142 & 0.173 & 0.176 & 0.177 & 0.180 & 0.180 & \textbf{0.182} \\
\hline
\end{tabular*}
\caption{Performance comparison between our method, the Datacomp baseline, and other participants' approaches.}
\label{tab:performance_all}
\end{table*}
\textit{Existing Baselines.}
Several SOTA methods have previously set benchmarks in data filtering. LAION and CLIP Score utilize the CLIP model to refine datasets, while Datacomp Filtering employs heuristic unimodal operators for targeted data refinement \cite{gadre2023datacomp}. The HYPerbolic Entailment (HYPE) Filtering technique\cite{kim2024hype} enhances data quality by integrating unimodal specificity with cross-modal alignment. LINE's strategy leverages large models for web data curation \cite{yokoo2023leveraging}. The Text-Masking and Re-Scoring (T-MARS) method corrects imbalances where textual features overpower visual ones \cite{maini2023t}, and the University of Wisconsin-Madison’s (WS) approach utilizes an ensemble of object detection methods to optimize data filtering \cite{huang2024multimodal}.

\textit{Performance Comparison.}
Building upon these foundations, EcoDatum enhances both efficiency and model training outcomes. As outlined in Table \ref{tab:performance_all}, using only 3.5 million data pairs from the original 12.8 million, EcoDatum achieved the highest average score of 0.182. This surpasses the performance of established methods like T-MARS and WS, both of which scored 0.180 across 38 diverse evaluation datasets. This curation strategy not only reduces computational overhead by 72\% but also significantly improves data quality. EcoDatum exceeds the ``No Filtering'' baseline score of 0.132 and the Datacomp Basic filtering score of 0.142 by 28\%. The integration of advanced methodologies like our optimized LabelModel for labeling functions tuning further refines the data curation process, setting new benchmarks in multimodal applications. The empirical results robustly validate our hypothesis that smaller, well-curated datasets can outperform larger, unfiltered datasets, underscoring the effectiveness of EcoDatum. Moreover, additional experiments show that EcoDatum consistently improves performance and scales effectively with increasing dataset size.

\begin{figure}[h]
    \centering
    \includegraphics[width=0.98\linewidth]{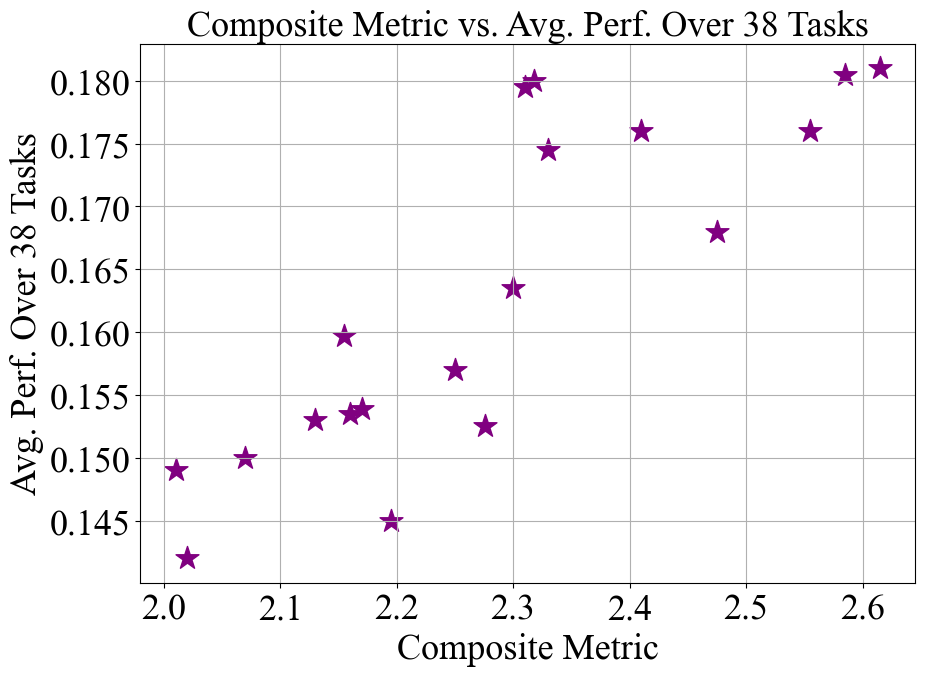}
    \caption{Composite Metric Validation with Repeated Experimental Downtasks Evaluations. The positive correlation indicates its capability to guide the tuning of the process.}
    \label{fig:metric}
\end{figure}

In this study, we introduce a composite metric designed to automatically optimize the generation of labeling functions (LFs), thereby facilitating the creation of a more accurate weak supervision label matrix. This optimization directly enhances the learning efficiency of the LabelModel, significantly improving its ability to assess data quality. To validate the effectiveness of this composite metric, we conducted a rigorous experimental case study. The process involved documenting a systematic search to identify the most effective LF combinations and repeatedly evaluating their impact on the average performance across a diverse set of 38 benchmark tasks. The results, depicted in Figure \ref{fig:metric}, demonstrate a consistent positive correlation between the composite metric scores and the model’s performance, affirming the metric’s utility in refining the data curation process.

\subsection{Ablation Study}

This experiment conducts a systematic evaluation of data filtering techniques to assess impacts on the performance of the deep learning model, as detailed in Table \ref{tab:performance_comparison_2}. The ``No Filtering'' condition acts as the control group.``Random Deduplication'' utilizes a stochastic method to eliminate duplicates, indicating that even indiscriminate reductions can improve model performance by balancing feature distribution.

\begin{table}[!ht]

\centering
\setlength{\tabcolsep}{3pt} 
\begin{tabular}{lccc}
\hline
\textbf{Methods} & \textbf{Dataset Size} & \textbf{Avg. Perf.} \\ 
\hline
No Filtering & 12.8M & 0.132 \\
\hline
Random Dedup. & 8.8M & 0.145 \\
Quality-Guided Dedup.(QGD) & 8.8M & 0.147 \\ 
\hline
QGD+Ens.(Uni.) & 3.5M & 0.154 \\ 
QGD+Ens.(Mul.) & 3.5M & 0.164 \\ 
\hline
QGD+Ens.(Uni.\&Global-Mul.) & 3.5M & 0.168 \\ 
QGD+Ens.(Uni.\&Local-Mul.)  & 3.5M & 0.155 \\ 
\hline
\textbf{Best Perf.} \\ QGD+Ens.(Uni.\&Mul.) & 3.5M & \textbf{0.182} \\ 
\hline
\end{tabular}
\caption{Performance comparison of different data curation and ensemble techniques over 38 downtasks.}
\label{tab:performance_comparison_2}
\end{table}

The introduction of QGD achieves a 1.4\% improvement over the random method with the same dataset size. Incorporating a unimodal operators' ensemble within the QGD framework results in a 4.8\% improvement, while a multimodal operators' ensemble leads to a more substantial 9.5\% enhancement. These results highlight the efficacy of both unimodal and multimodal operator ensembles in data curation. By integrating QGD with both unimodal and multimodal ensembles, the combined approach outperforms all others, showing a 45.4\% improvement in performance compared to the ``No Filtering" baseline. These experiments illustrate that EcoDatum strategically integrates advanced deduplication techniques and sophisticated ensemble frameworks to markedly elevate data quality, optimizing the pretraining process for multimodal models.

We conduct another ablation study to assess the individual contributions of data processing operators in data curation. By applying each operator independently and incrementally adding them, we explored their impact on downstream tasks. This approach allowed us to identify the most effective combinations of operators, significantly streamlining the optimization process. Through meticulous integration and refinement of labeling function (LF) constructions, we determined the most efficient operator combinations, thereby enhancing the accuracy and efficacy of our data curation methods. This conclusion suggests a strategic approach when dealing with massive web data and limited computational resources: focusing on alignment techniques can lead to more efficient data filtering. Such a focus can improve the generalization performance of multimodal models. Potentially, this experiment could pave the way for future research, indicating that more advanced image-text matching techniques might result in even better multimodal curation outcomes.

\vspace{-0.2cm}
\section{Conclusion and Future Work}
The volume of web-crawled datasets is rapidly expanding, and training multimodal models with such data are increasingly prevalent. This paper addresses the challenge of variable sample quality in web-crawled datasets by introducing a novel data curation framework, EcoDatum, designed to select high-quality data. EcoDatum begins with quality-guided deduplication to preprocess the data, followed by the integration of unimodal and multimodal operators into a weak supervision ensemble model, LabelModel, and have employed a search-based optimization method to refine the labeling matrix within LabelModel. Our experiments demonstrate robust performance across all evaluated tasks, securing a 1\textsuperscript{st} place ranking in the small-scale track of the DataComp benchmark. While this study validated EcoDatum on a small dataset, future work will extend the evaluation to larger datasets. This expansion will further test the scalability of EcoDatum, aiming to solidify its effectiveness and efficiency in enhancing the training of multimodal models with diverse, large-scale web-crawled data.

\section{Acknowledgments}
This work was supported in part by China NSF grant No. 62202297 and in part by the Open Project Program of the Laboratory of Pinghu. The opinions, findings, conclusions, and recommendations expressed in this paper are those of the authors and do not necessarily reflect the views of the funding agencies or the government.

\bibliography{aaai25}

\newpage

\section{A.Appendix}

\subsubsection{A.1.Exploring the balance between quantity and performance.}

Approximately the top 40\% samples by the EcoDatum quality score after deduplication (around 3.5M) provides the best balance between quality and quantity,

\begin{figure}[h]
    \centering
    \includegraphics[width=1\linewidth]{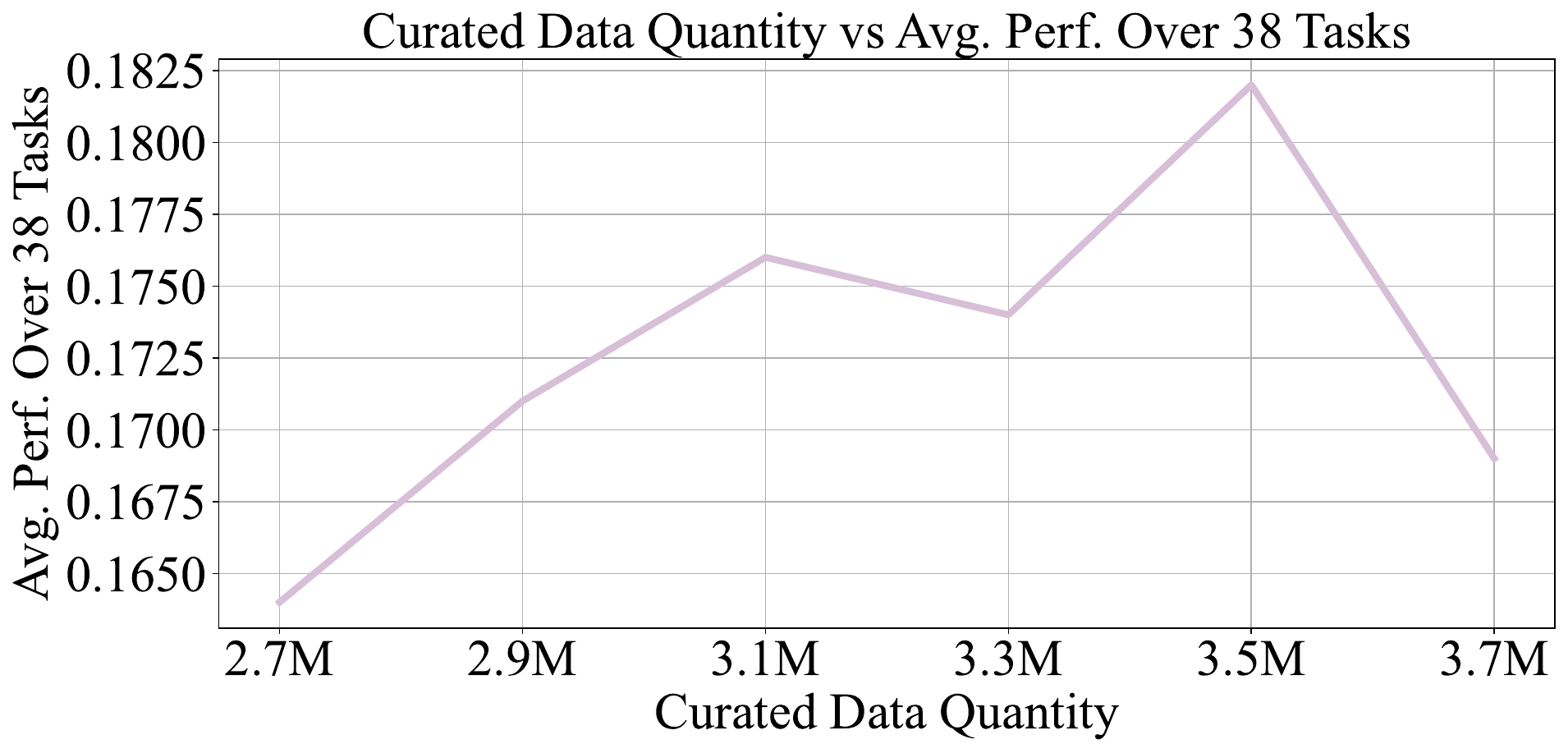}
    \caption{Vary the filtered dataset quantity and compare the results of repeated experiments to find an optimal balance at ~3.5M samples under EcoDatum curation sorting.}
    \label{fig:quant}
\end{figure}

\subsubsection{A.2.Single Operator Curations' Results.}

We conduct single-operator filtering experiments on the entire dataset to better utilize data curation operators and design a more efficient integration. In these experiments, we do not standardize the amount of data selected by each operator. Instead, to maximize the reflection of each operator's unique characteristics, we use their default settings for data selection (partially referencing related works, such as DataComp).

\begin{table}[!h]
\centering
\begin{tabular}{lccc}
\hline
\textbf{Methods} & \textbf{Dataset Size} & \textbf{Avg. Perf.} \\ 
\hline
No Filtering &  12.8M & 0.132 \\ 
\hline
Image-Blurry & 10.8M & 0.138 \\
Image-Geometric & 10.7M & 0.144 \\
Text-Language Id. & 5.3M & 0.150 \\
Text-ICC & 3.9M & 0.147 \\
\hline
CLIP & 7.5M& 0.146 \\
H-CLIP & 4.3M & 0.160 \\
V-CLIP & 7.2M & 0.154 \\
GroundingDINO & 6.2M & 0.141 \\
\hline
\end{tabular}
\caption{Performance comparison of single operator curation.}
\label{tab:single_operator_performance}
\end{table}

The results shown in Table \ref{tab:single_operator_performance} demonstrate that the global feature alignment operator achieved the best single-operator performance on the same benchmark. This aligns with the findings from the progressive integration experiments discussed in the main text, reinforcing the conclusion that the global feature alignment operator makes a significant contribution to image-text data filtering tasks.

Additionally, by using a single operator independently, we observed that data curation with just one operator introduces certain biases. This bias arises because each operator evaluates data samples from a different perspective. Therefore, this finding highlights the importance and value of effectively utilizing multiple operators in a complementary manner.

\subsubsection{A.3.Quality Guided Deduplication (QGD) Visualization.}

\begin{figure}[ht!]
    \centering
    \includegraphics[width=1\linewidth]{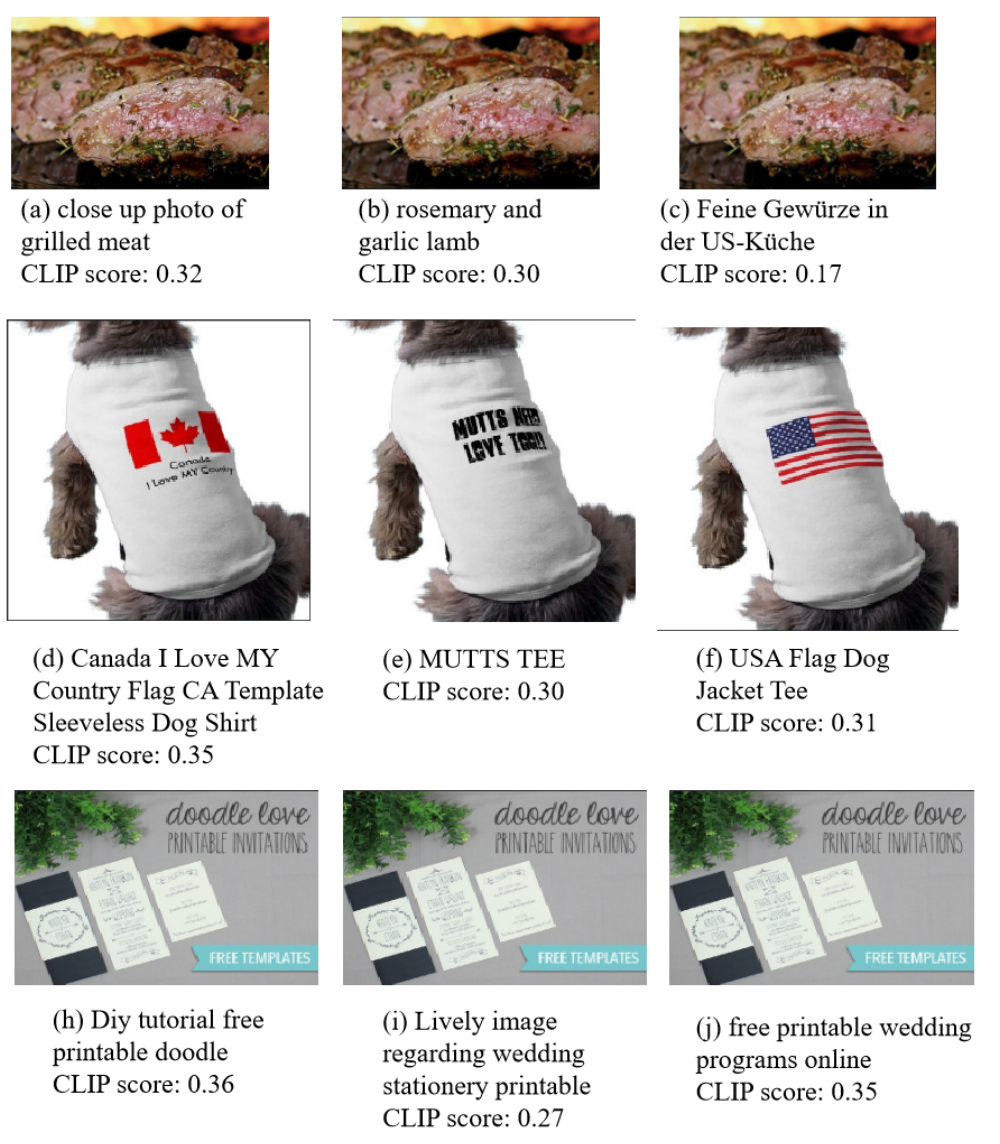}
    \caption{Quality Guided Deduplication Samples: Among each duplicated group, only the finely aligned image-text pairs are kept and left others to improve the overall distribution with maintaining cross-modal quality.}
    \label{fig:dedup_appendix}
\end{figure}

In the context of multimodal image-text datasets, traditional unimodal deduplication methods that randomly remove duplicates are insufficient. These unimodal methods often overlook the crucial interplay between image and text, leading to potential misalignments in retained data. Random selection introduces uncertainty regarding the degree of coherence between the two modalities, which can compromise the multimodal dataset's utility.

In the EcoDatum approach, deduplication takes into account the quality of image-text alignment. For duplicate image-text pairs, as shown in Figure \ref{fig:dedup_appendix}, rather than randomly selecting which to keep, the process identifies and retains the pair with the best global alignment. This method ensures that the highest quality and most relevant examples are preserved. It significantly improves the overall data quality and distribution, enhancing the dataset’s diversity and reducing redundancy without sacrificing alignment accuracy.

QGD is particularly beneficial because it:
\begin{itemize}
    \item \textit{Preserves Semantic Integrity:} Ensure that the retained data is the most representative in terms of both visual and textual accuracy.
    \item \textit{Enhances Model Training:} High-quality, well-aligned data leads to more accurate and robust models.
    \item \textit{Maintains Dataset Diversity:} By selectively keeping the best-aligned pairs, the method supports a diverse and balanced dataset, which is critical for generalizability.
\end{itemize}

Thus, moving away from random deduplication to a more deliberate, quality-guided approach significantly bolsters the effectiveness and reliability of multimodal datasets.

\subsubsection{A.4.Comparison With Different Global Feature Alignment Strategies.}

For certain horizontally symmetrical letters, simple horizontal flipping may fail to distinguish them, leading to mismatched image-text pairs being incorrectly judged as matching, thereby introducing noise. The global feature alignment operator in EcoDatum addresses this by performing flips in both directions, ensuring that the impact of symmetrical text in images on image-text similarity calculations is fully considered.

\begin{figure}
    \centering
    \includegraphics[width=1\linewidth]{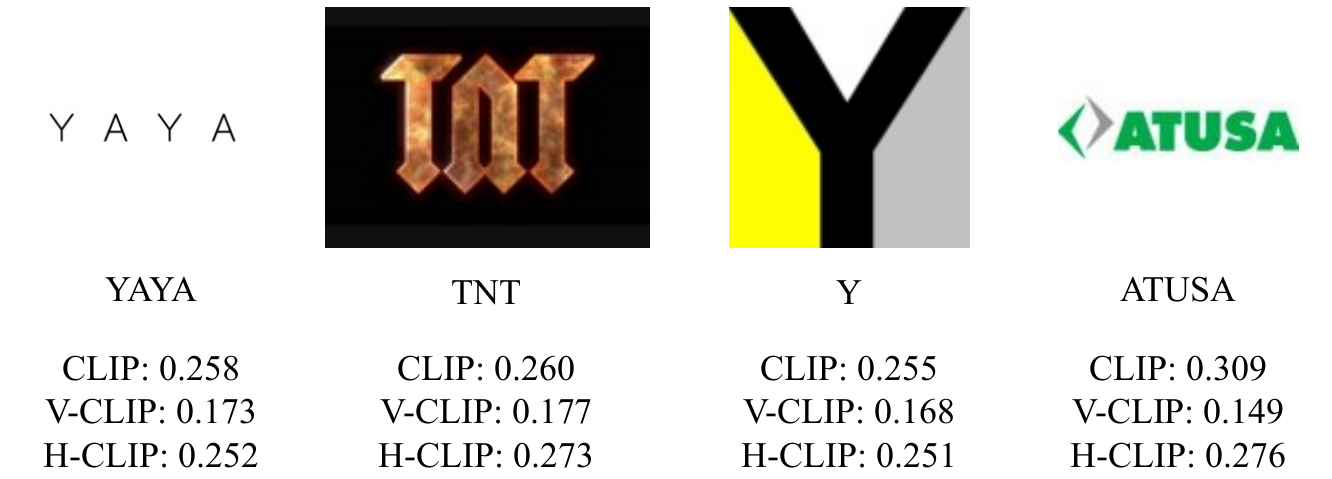}
    \caption{Hard Bad Cases Mined by V-CLIP Operator}
    \label{fig:3clip}
\end{figure}

Through visualized results as Figure \ref{fig:3clip}, it was observed that for images containing symmetrical letters, even when both CLIP and H-CLIP fail to accurately assess the quality, the introduction of V-CLIP still provides low scores, indicating lower data quality. Additionally, extended experiments in Table \ref{tab:3clip} showed that removing V-CLIP led to a performance decline, further demonstrating that augmentation in both directions is meaningful and beneficial for maintaining robust data quality assessment.

\begin{table*}[h!]
    \centering
    \begin{tabular}{cccccccccc}
        \hline
         & \textbf{Blurry} & \textbf{Geometry} & \textbf{Language} & \textbf{ICC} & \textbf{CLIP} & \textbf{H-CLIP} & \textbf{V-CLIP} & \textbf{GDINO} & \textbf{Perf.} \\
        \hline
        Control Group. & 0.5593 & 0.5341 & 0.8438  & 0.7467 & 0.7334 & - & 0.5875 & 0.7467 & 0.169 \\
        \hline
        Best Perf. & 0.7564 & 0.6186 & 0.6595  & 0.5832  & 0.8078  & 0.9154 & 0.8775 & 0.6329 & \textbf{0.182}  \\
        \hline
    \end{tabular}
    \caption{Exp. (Without H-CLIP) vs. Best Perf. (With H-CLIP)}
    \label{tab:3clip}
\end{table*}

\subsubsection{A.5.Local Feature Alignment Visualizations.}

\begin{figure}[h!]
    \centering
    \includegraphics[width=1\linewidth]{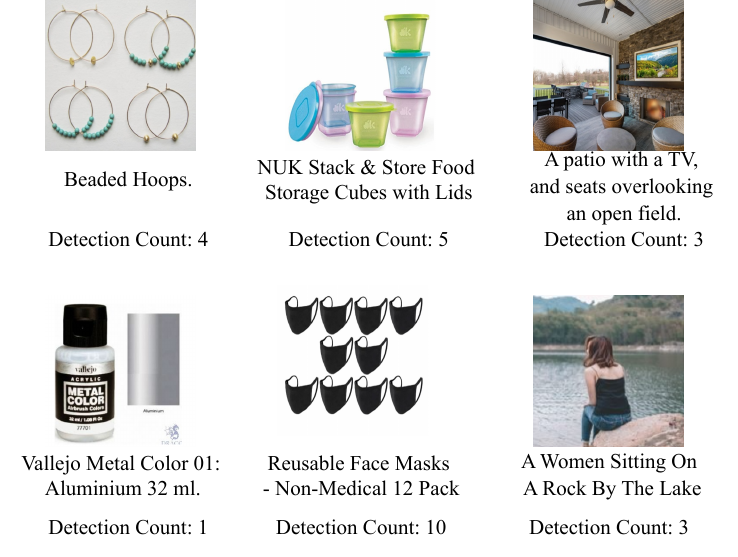}
    \caption{High alignment quality data detected by the Local Cross-Modal Feature Alignment Operator}
    \label{fig:gdinogood}
\end{figure}

\begin{figure}[h!]
    \centering
    \includegraphics[width=1\linewidth]{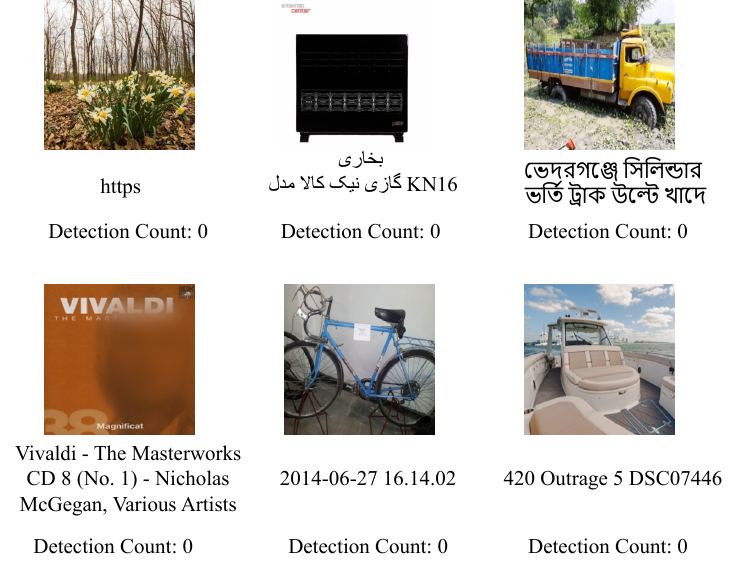}
    \caption{Low alignment quality data detected by the Local Cross-Modal Feature Alignment Operator}
    \label{fig:gdinobad}
\end{figure}

Utilizing the GroundingDINO model, EcoDatum enhances the precision of aligning text descriptions with their corresponding visual content within the dataset. This model effectively integrates and cross-verifies textual and visual data, employing a quantitative metric based on the count of bounding boxes whose confidence scores exceed a predefined threshold. This approach is illustrated in Figures \ref{fig:gdinogood} and \ref{fig:gdinobad}, showcasing examples of effective and poor data alignments.

Effective Alignments (Figure \ref{fig:gdinogood}): High detection counts, as seen with items like "NUK Stack \& Store Food Storage Cubes" and "Reusable Face Masks," indicate robust correspondence between the text descriptions and visual elements. These examples highlight the model's capability to retain rich, descriptive multimodal content, valuable for training and analytical applications.

Poor Alignments (Figure \ref{fig:gdinobad}): Conversely, detection counts of zero in examples such as "A Truck on the grass" and a "bicycle parked outside" demonstrate the model's critical role in identifying mismatches. These instances occur when textual descriptions do not visually correspond to the images, or when the described objects are absent from the visual content. Such discrepancies point to insufficient or inaccurate text descriptions, where the text either describes non-existent objects or is too vague.

By highlighting these mismatches, GroundingDINO ensures the dataset excludes low-quality entries where textual and visual data do not align, thereby maintaining only content with verified and accurate multimodal relationships. This selective curation is essential for developing datasets that are not only diverse but also consistently reliable for training robust multimodal systems.

\subsubsection{A.6.Data Curation Operators Ensemble's Analysis.}

Here, we documented and analyzed the attributes of each Labeling Function and explored the operator weights modeled by the LabelModel during the integration learning phase, as shown in Table \ref{tab:function_metrics}. It was observed that these operator weights correspond with the earlier single-operator results, indicating a higher reliance on the global alignment operator. This further corroborates the importance of cross-modal global feature alignment in image-text data curation.

\begin{table*}[h!]
    \centering
    \begin{tabular}{lcccc}
        \hline
        \textbf{Operators} & \textbf{Coverage} & \textbf{Overlaps} & \textbf{Conflicts} & \textbf{Weights} \\ 
        \hline
        Blurry & 0.003 & 0.003 & 0.002 & 0.756 \\ 
        Geometry & 0.014 & 0.014 & 0.011 & 0.619 \\ 
        \hline
        Language Identification & 0.532 & 0.524 & 0.362 & 0.660 \\ 
        ICC & 0.345 & 0.343 & 0.239 & 0.583 \\ 
        \hline
        CLIP& 0.739 & 0.733 & 0.471 & 0.808 \\ 
        H-CLIP& 0.754 & 0.738 & 0.410 & 0.915 \\ 
        V-CLIP& 0.658 & 0.657 & 0.409 & 0.878 \\ 
        \hline
        GroundingDINO& 0.423 & 0.421 & 0.312 & 0.633\\ 
        \hline
    \end{tabular}
    \caption{Ensembled Operators Attributes and Analyzed Weights}
    \label{tab:function_metrics}
\end{table*}

\subsubsection{A.7.Search-based Optimization Implementation Details.}

\begin{figure}[h]
    \centering
    \includegraphics[width=1\linewidth]{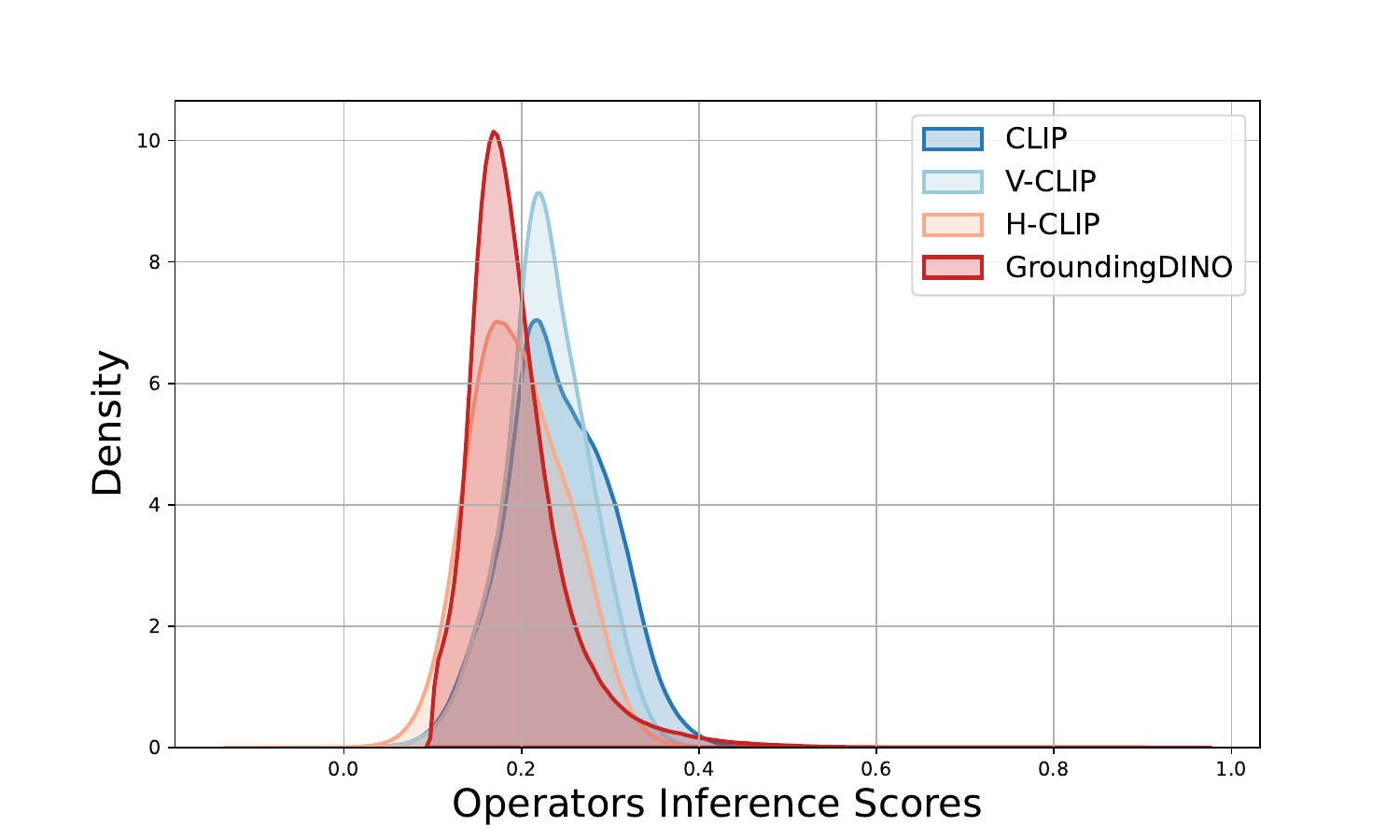}
    \caption{Multimodal Operators Inf. Scores' Distributions}
    \label{fig:mm_scores}
\end{figure}

\begin{figure}[h!]
    \centering
    \includegraphics[width=1\linewidth]{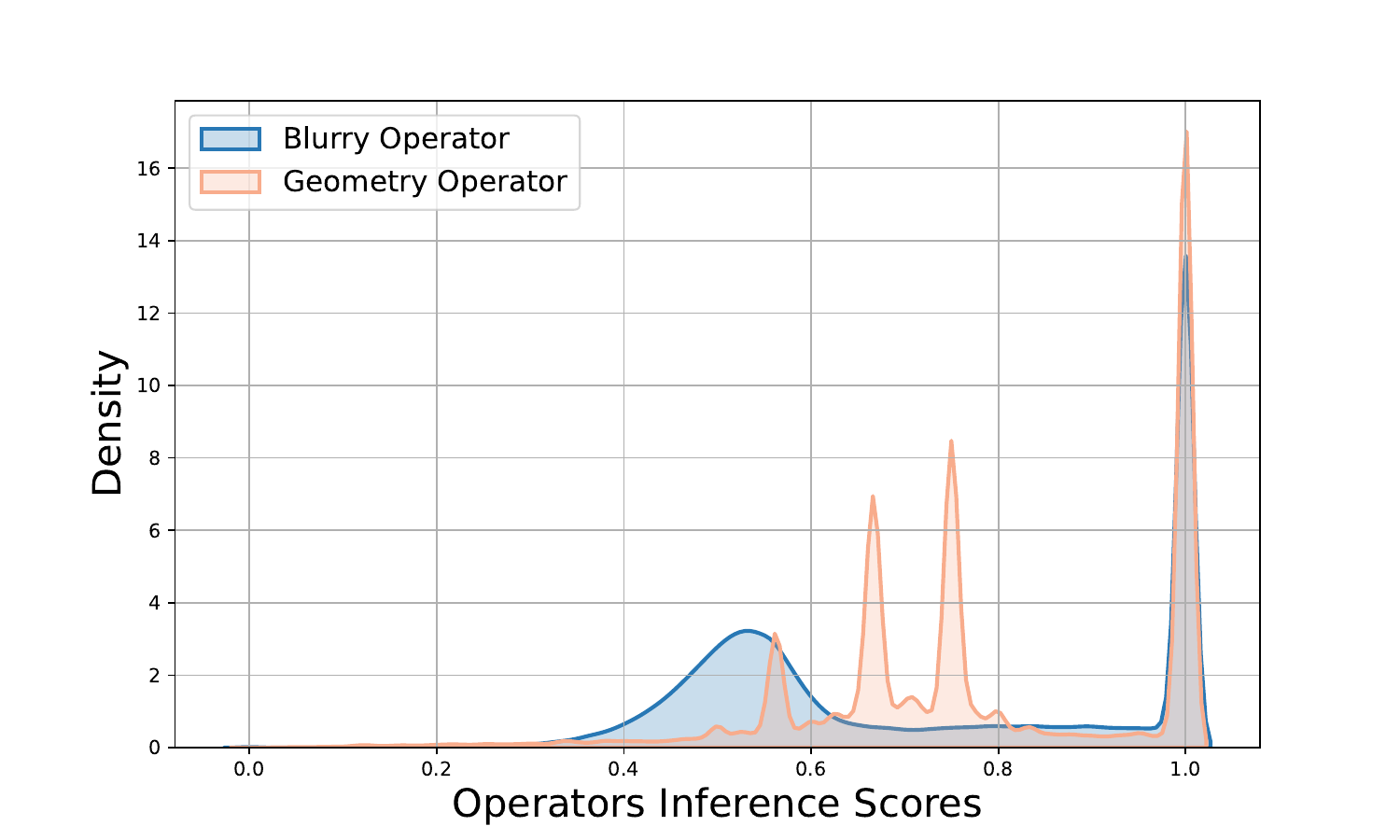}
    \caption{Unimodal(Visual-based) Operators Inf. Scores' Distributions}
    \label{fig:img_dist}
\end{figure}

\begin{figure}[h!]
    \centering
    \includegraphics[width=1\linewidth]{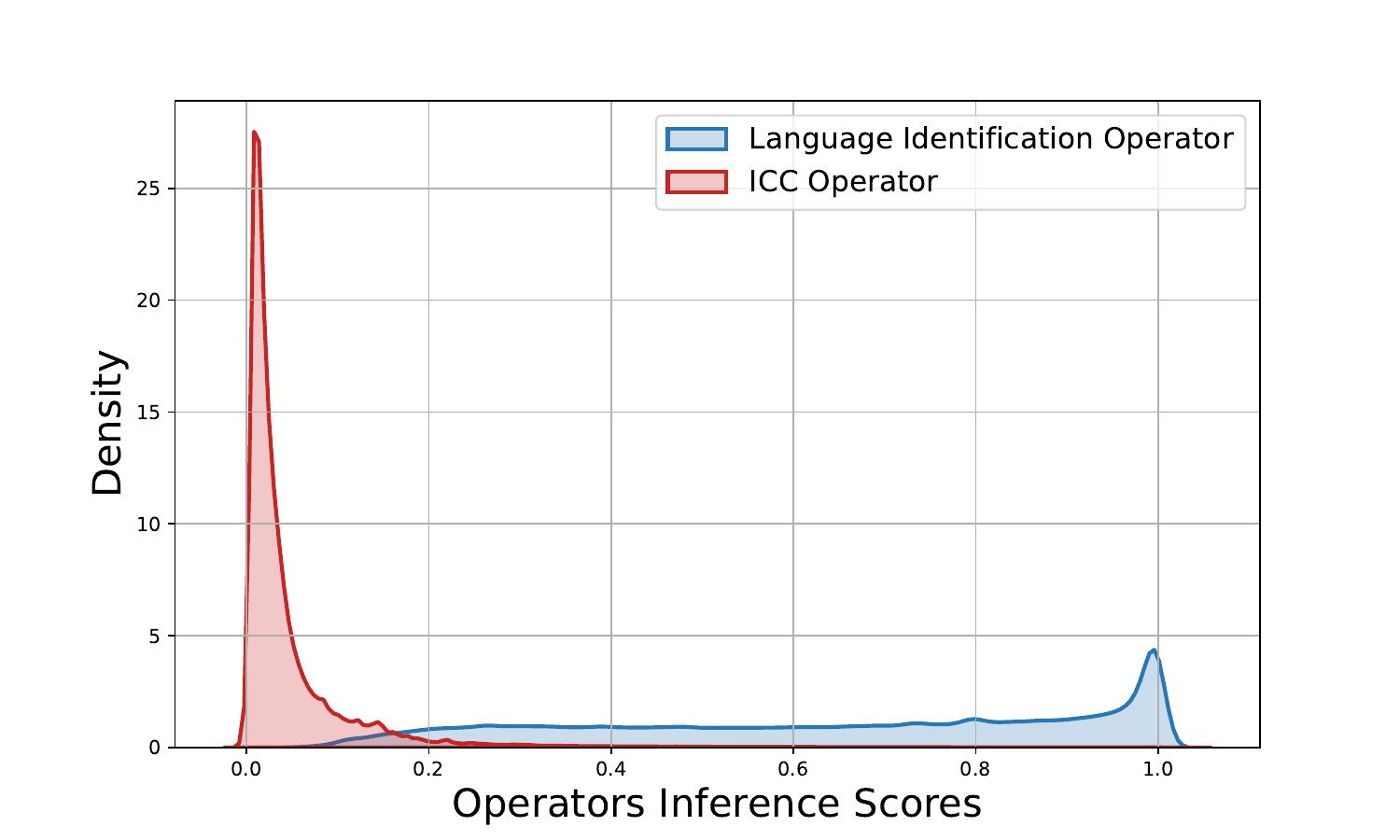}
    \caption{Unimodal(Textual-based) Operators Inf. Scores' Distributions}
    \label{fig:txt_dist}
\end{figure}

To better configure the candidate Labeling Functions, we first analyze the inference score distribution of each operator across the entire dataset, as shown in Figure \ref{fig:mm_scores}, Figure \ref{fig:img_dist} and Figure \ref{fig:txt_dist}. We then introduced the TopK values to construct candidate Labeling Functions to assist in converting output scores into weak supervision labels, as shown in Table \ref{table:performance_metrics}. 

\begin{table*}[h!]
\centering
\begin{tabular}{l|c|c|c}
\hline
\textbf{TopK} & \textbf{Global Feature Alignment} & \textbf{ICC Operator} & \textbf{Local Feature Alignment} \\ 
\hline
Top20 & 0.271 & 0.064 & 0.225 \\
Top30 & 0.252 & 0.041 & 0.206 \\
Top40 & 0.235 & 0.035 & 0.192 \\
Top50 & 0.220 & 0.023 & 0.180 \\
Top60 & 0.211 & 0.017 & 0.169 \\
Top70 & 0.192 & 0.013 & 0.158 \\
Top80 & 0.187 & 0.010 & 0.146 \\ \hline
\end{tabular}
\caption{Single Operators Inference TopK Scores}
\label{table:performance_metrics}
\end{table*}

The final experiments confirmed that our proposed EcoDatum method, which currently achieves SOTA on the DataComp benchmark, can identify the most suitable configurations using the proposed search optimization method through this well-constructed candidate pool.

\subsubsection{A.8.Search-based Optimation Validation.}

To validate the automatic optimization method used for Labeling Functions in EcoDatum, we randomly initialize the weights for three sets of LabelModels. We then conducted repeated experiments by integrating the same unimodal or multimodal operators while applying these random weights, and we retained the previously validated optimal data quantity. 

\begin{table*}[h!]
    \centering
    \begin{tabular}{c|ccccccccc}
        \hline
        \textbf{Settings}
         & \textbf{Blurry} & \textbf{Geometry} & \textbf{Language} & \textbf{ICC} & \textbf{CLIP} & \textbf{H-CLIP} & \textbf{V-CLIP} & \textbf{GDINO} & \textbf{Perf.} \\
        \hline
        Exp1 & 0.5924 & 0.1241 &0.6004  & 0.9877 & 0.3738 & 0.5904 & 0.9737 & 0.7416 & 0.157 \\
        \hline
        Exp2 & 0.8402 & 0.1481 &  0.3681& 0.0736 & 0.2766 & 0.8365 & 0.9522 & 0.0082 & 0.141 \\
        \hline
        Exp3 & 0.0827 & 0.7436 & 0.4301  &0.8606  &0.7251  & 0.8246 & 0.8109 & 0.7574 & 0.161  \\
        \hline
        Best Perf. & 0.7564 & 0.6186 & 0.6595  & 0.5832  & 0.8078  & 0.9154 & 0.8775 & 0.6329 & \textbf{0.182}  \\
        \hline
    \end{tabular}

    \caption{Random weights-assigned ensembled curation comparisons, here exp1, exp2, exp3 take three different LabelModel random weights combinations.}
    \label{table:random_1}
\end{table*}

As demonstrated in Table \ref{table:random_1}, the results showed that random initialization of weights not only failed to produce excellent integration performance for the Data Curation task, but also led to negative results in Experiment Group 2. This indicates that without proper integration of the operators, the LabelModel is ineffective for data quality assessment and cannot adequately manage the data.

\subsection{A.9. Scaling Data Experiments}

We conducted additional experiments on EcoDatum scalability on datasets of different sizes. The results demonstrate that EcoDatum consistently enhances performance across different dataset scales, indicating both robustness and scalability of the proposed method. The detailed performance metrics are summarized in the Table \ref{tab:scalability}.

From these experiments, it is evident that the performance improvement achieved by EcoDatum becomes more pronounced as the dataset size increases. This trend suggests that the method is well-suited for scaling and can deliver greater benefits in larger datasets. In addition, it is worth noting that state-of-the-art visual-language models (VLMs) have achieved remarkable performance using smaller but well-curated datasets. For instance, LLaVA-1.5-HD uses only 558K images for pretraining and 665K for fine-tuning, achieving an 81.8 score on VQAv2. In contrast, Qwen-VL uses 1.4B images for pretraining and 50M for fine-tuning but attains a lower score of 78.8. These examples highlight that high-quality data curation can outweigh the advantages of larger but noisier datasets. This aligns with EcoDatum's central premise: emphasizing data quality over quantity can lead to significant performance improvements.

\begin{table}[h]
\centering
\begin{tabular}{|>{\raggedright}p{3cm}|c|c|c|} 
\hline
\textbf{Dataset Size} & \textbf{0.128m} & \textbf{1.28m}  & \textbf{12.8m}  \\ \hline
No-filtering          & 0.073           & 0.088           & 0.132            \\ \hline
EcoDatum              & 0.075           & 0.104           & 0.182            \\ \hline
Improvement           & 0.002           & 0.016           & 0.050            \\ \hline
\end{tabular}
\caption{Performance of EcoDatum Across Different Dataset Sizes} 
\label{tab:scalability}
\end{table}

\subsection{A.10.Selected or Filtered Subsets Visualization.}

\begin{figure*}[h]
    \centering
    \includegraphics[width=1\linewidth]{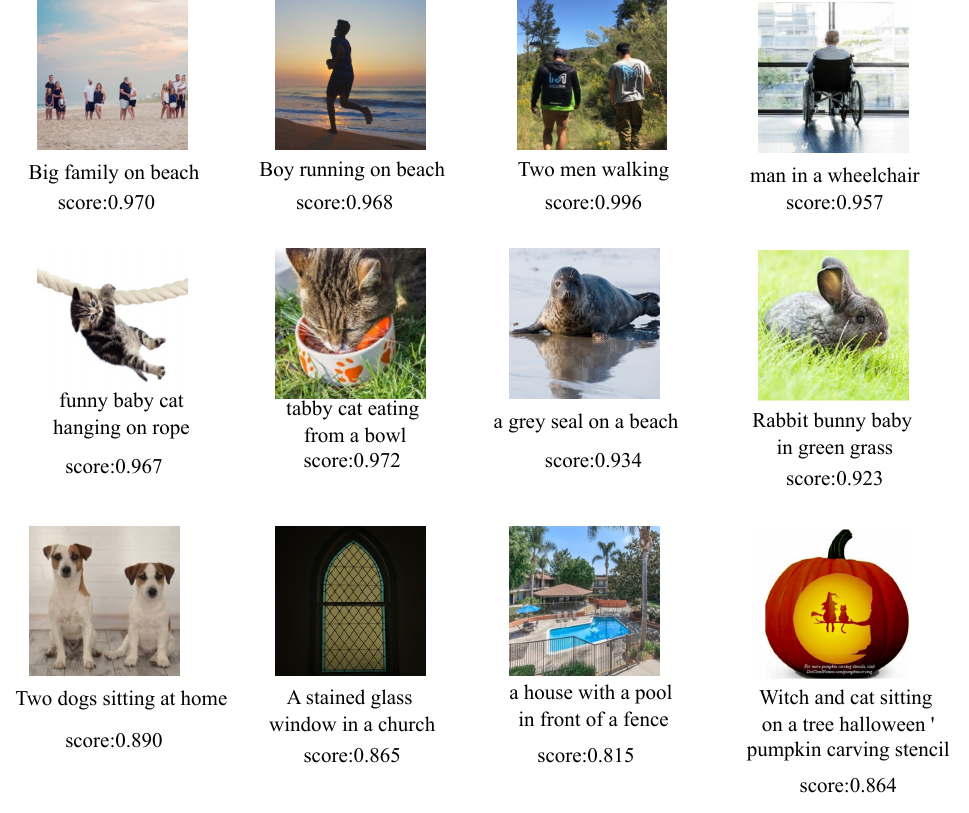}
    \caption{Selected ``High Quality'' Data Samples from EcoDatum. ``High Quality'' image-text data refers to pairs where the content is clear, accurate, and highly consistent. In these pairs, the text description precisely reflects the image content, with strong relevance and rich information between the two. The annotations are well-structured and suitable for various tasks, thereby effectively enhancing the comprehension and generalization capabilities of multimodal models.}
    \label{fig:goodsamples}
\end{figure*}

\begin{figure*}[h]
    \centering
    \includegraphics[width=1\linewidth]{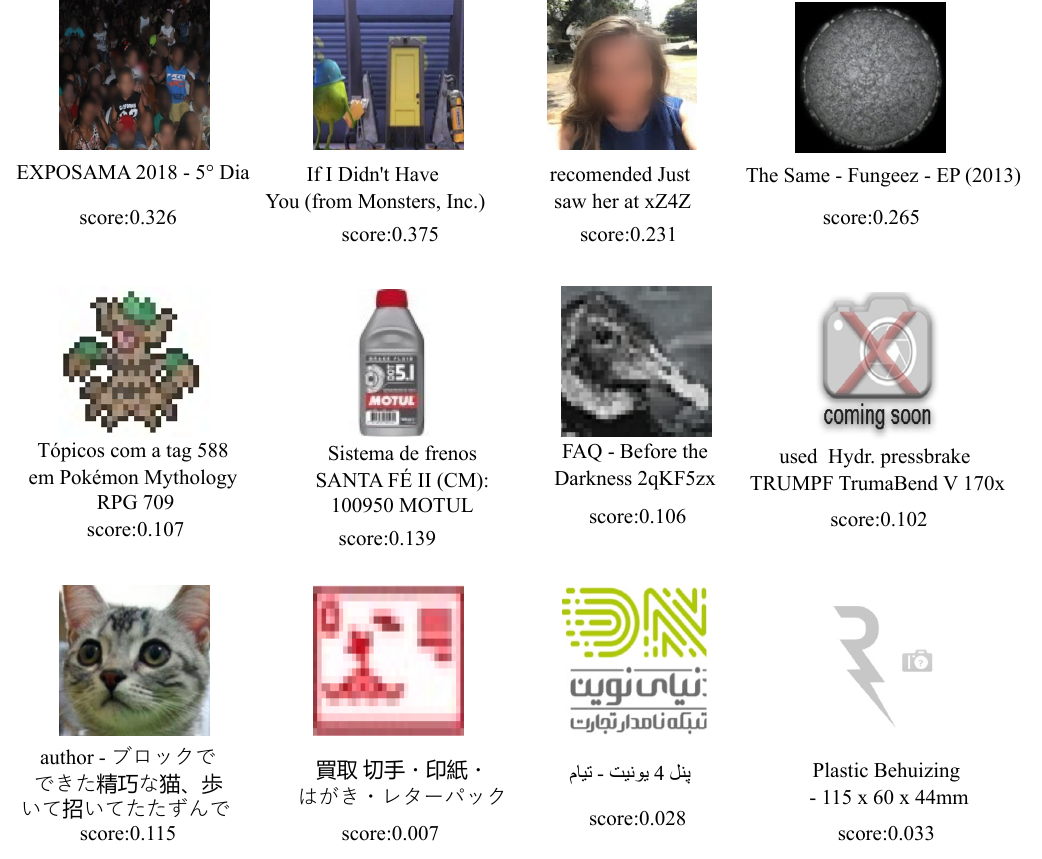}
    \caption{Filtered ``Low Quality'' Data Samples from EcoDatum. ``Low-quality'' image-text data typically manifests as blurry or low-resolution images, text with spelling and grammatical errors, mismatched content between images and text, and insufficient or redundant information. Training models with such data can lead to decreased performance, increased misclassification, and ultimately compromise the accuracy and reliability of the model in real-world applications.}
    \label{fig:badsamples}
\end{figure*}

Figures \ref{fig:goodsamples} and \ref{fig:badsamples} illustrate the significant differences in quality between selected high-quality data samples and filtered low-quality data samples by our data curation framework, EcoDatum. These visualizations showcase the effectiveness of the EcoDatum in distinguishing and retaining highly aligned image-text pairs, while identifying and excluding low-quality data. The high-quality subset in Figure \ref{fig:goodsamples} features images with clear visuals and precise text descriptions that are semantically rich, enhancing the potential of training more effective multimodal models. Conversely, Figure \ref{fig:badsamples} displays low-quality samples characterized by poor image resolution, misaligned texts, or irrelevant content, underscoring the importance of rigorous data filtering to maintain the integrity and utility of the dataset.

\end{document}